# Advancements and limitations of LLMs in replicating human color-word associations

Makoto Fukushima[1]*, Shusuke Eshita[2], Hiroshige Fukuhara[2]

[1] Deloitte Analytics R&D, Deloitte Tohmatsu Risk Advisory LLC, Tokyo, Japan. [2] Sony Design Consulting Inc., Tokyo, Japan

Corresponding author: Makoto Fukushima (e-mail: makoto.fukushima@tohmatsu.co.jp).

## ABSTRACT

Color-word associations play a fundamental role in human cognition and design applications. Large Language Models (LLMs) have become widely available and have demonstrated intelligent behaviors in various benchmarks with natural conversation skills. However, their ability to replicate human color-word associations remains understudied. We compared multiple generations of LLMs (from GPT-3 to GPT-4o) against human color-word associations using data collected from over 10,000 Japanese participants, involving 17 colors and 80 words (10 word from eight categories) in Japanese. Our findings reveal a clear progression in LLM performance across generations, with GPT-4o achieving the highest accuracy in predicting the best voted word for each color and category. However, the highest median performance was approximately 50% even for GPT-4o with visual inputs (chance level of 10%). Moreover, we found performance variations across word categories and colors: while LLMs tended to excel in categories such as Rhythm and Landscape, they struggled with categories such as Emotions. Interestingly, color discrimination ability estimated from our color-word association data showed high correlation with human color discrimination patterns, consistent with previous studies. Thus, despite reasonable alignment in basic color discrimination, humans and LLMs still diverge systematically in the words they assign to those colors. Our study highlights both the advancements in LLM capabilities and their persistent limitations, raising the possibility of systematic differences in semantic memory structures between humans and LLMs in representing color-word associations.

## Introduction

Recent advancements in Large Language Models (LLMs) have demonstrated their ability to interact with people quite naturally. This capability has reached a level where people can discern whether an LLM is their conversation partner only at chance level in certain conversational settings [1]. Such performance suggests LLMs' potential to replicate not only human responses in conversation but also the cognitive processes underlying them. This implies the possibility of using LLMs as substitutes for human participants in probing cognitive functions, which traditionally rely on human survey or performance data and are time-consuming and constrained by limited sample sizes. Utilizing LLMs could enable large-scale studies of the mechanisms underlying human cognition, which would not be feasible through data collection from human participants alone. Previous studies have explored the cognitive capabilities of LLMs, indicating that they can approximate human cognition, albeit with varying success across different LLM iterations [2,3].

Color-word associations play an important role in human cognition and are fundamental to various fields, including psychology, linguistics, and design [4]. These associations, rooted in perceptual and cognitive processes, have been the subject of extensive research, with studies exploring their universal and culture-specific aspects [5–7]. In a practical context, professional design requires not only creating representations that are aesthetically pleasing but also achieving functional meaning. Therefore, when designing color schemes, it is important to identify the information associated with colors to create expressions that convey the intended functional information. Previous studies have developed valuable tools for applying color-word relationships in design contexts based on human responses [8]. On the other hand, recent studies have explored data-driven approaches to capturing these associations, leveraging web-based data and machine learning techniques. One such approach adapts natural language processing techniques to create embeddings for both words and colors based on web-sourced images [9].

Several recent studies have begun to probe color knowledge in large language models: GPT-3 has been shown to generate basic color terms at frequencies comparable to those produced by human non-synaesthetes [10], and newer models such as GPT-4 display color-discrimination abilities that correlate well with human judgements [11, 12]. These studies demonstrated the potential of LLMs to replicate certain aspects of human color perception. Yet, there has been little study directly evaluating LLMs' capability to replicate human color-word associations. Thus, our study here focuses on assessing the ability of LLMs to replicate human color-word associations. Building upon our prior work that analyzed responses from over 10,000 Japanese participants [7], we evaluate LLMs' performance in matching these human color-word associations across a range of LLM versions from GPT-3 to GPT-4o, with both text and visual color inputs.

## Material and Methods

### A. Human Data Collection

The study utilized anonymized data from a web-based survey conducted by a professional survey company, involving 12,369 Japanese participants. Participants had consented to research participation by providing written informed consent before completing the questionnaire. The sample was selected to represent the demographic distribution of gender, age, and residential areas in Japan. A preliminary analysis of these data has previously reported elsewhere [7,21]. The participants in this study were not screened for normal color vision. However, a previous survey by the same company found that approximately 0.2% of participants had atypical color vision. Participants completed a multiple-choice questionnaire, selecting words associated with each of 17 colors (Black, Burgundy, Navy, Dark Green, Orange, Violet, Red, Blue, Green, Yellow, Purple, Salmon Pink, Greyish Blue, Light Green, Cream, Light Purple, White; Table 1) across eight categories: Taste, Impression, Emotion, Field, Scenery, Product/Service, Landscape, and Rhythm. Each category offered ten-word choices (Table 2). The survey consisted of 153 questions (17 colors × 9 categories, including

| Color Name | RGB Value | |
|---|---|---|
| Black | [0, 0, 0] |  |
| Burgundy | [123, 0, 0] |  |
| Navy | [0, 14, 121] |  |
| Dark Green | [16, 55, 0] |  |
| Orange | [251, 184, 0] |  |
| Violet | [77, 16, 151] |  |
| Red | [246, 0, 0] |  |
| Blue | [0, 30, 255] |  |
| Green | [46, 169, 0] |  |
| Yellow | [255, 253, 0] |  |
| Purple | [144, 27, 255] |  |
| Salmon Pink | [230, 142, 210] |  |
| Greyish Blue | [142, 176, 210] |  |
| Light Green | [166, 201, 169] |  |
| Cream | [246, 240, 162] |  |
| Light Purple | [170, 158, 213] |  |
| White | [255, 255, 255] |  |

**Table 1: Color Names and Corresponding RGB Values.** This table presents a list of 17 colors used in the study, along with their corresponding RGB (Red, Green, Blue) values and image representations. RGB values were used as inputs for LLMs, and the visual representations were used as inputs for the vision models and human participants. The RGB values range from 0 to 255 for each component.

| Question | Answer Choices (10 words/options) |
|---|---|
| Please choose one Taste that you associate with this color from the following options. | Sweet (甘い), Sour (酸っぱい), Refreshing (爽やか), Astringent (渋い), Bitter (苦い), Delicious (美味しい), Spicy (辛い), Fragrant (香ばしい), Rich (濃厚), Salty (しょっぱい) |
| Please choose one Impression that you associate with this color from the following options. | Calm (静寂), Active (活発), Lovely (可愛い), Bright (明るい), Dark (暗い), Light (軽快), Heavy (重い), Youthful (若々しい), Antique (古風), Serious (真面目) |
| Please choose one Emotion that you associate with this color from the following options. | Sad (悲しい), Happy (楽しい), Angry (怒り), Lonely (寂しい), Uneasy (不安), Pride (高揚), Happiness (幸福), Envy (嫉妬), Harmony (和み), Empathy (共感) |
| Please choose one Field that you associate with this color from the following options. | Sports (スポーツ), Medicine (医療), Chemistry (化学), Religion (宗教), Literature (文学), Welfare (福祉), Agriculture (農業), Technology (技術), Arts (芸術), Industry (工業) |
| Please choose one Scenery that you associate with this color from the following options. | City (都市), Countryside (田園), Wilderness (荒野), Coast (海岸), Deep Sea (深海), Forest (森林), Grassland (草原), Hill (丘陵), Stream (清流), Desert (砂漠) |
| Please choose one Product that you associate with this color from the following options. | Automobile (自動車), Electronic Devices (電子機器), Internet (インターネット), Food (食品), Beverage (飲料), Clothing (服飾), Travel (旅行), Finance (金融), Construction (建築), Transportation (輸送) |
| Please choose one Landscape that you associate with this color from the following options. | Sunny (晴れ), Rainy (雨), Snowy (雪), Cloudy (曇り), Morning (早朝), Noon (昼), Evening (夕方), Night (夜), Gloomy (暗い), Cold (寒い) |
| Please choose one Rhythm that you associate with this color from the following options. | Slow (緩やか), Fast (早い), Light (軽快), Mechanical (機敏), Irregular (不規則), Simple (単調), Slow (遅い), Pure (鈍い), Repetitive (反復), Halting (静止) |

**Table 2: Questions and answer choices used in the color-word association survey.** Note that all questions and answers were presented in Japanese to both human participants and LLMs.

Shape that was not included in the current study). To reduce participant burden, respondents were divided into three balanced groups of 4,123 individuals (2,066 men and 2,057 women), with each answering 51 questions. The ten response words in each category were curated by two co-authors who are professional designers, selecting terms that capture distinctions useful in everyday design practice, and the 17-color palette was carried over unchanged from our earlier color-association survey [7] to maintain continuity with that benchmark.

*B. LLM Data Collection*

The following six LLM models were tested: gpt-4o-2024-05-13, gpt-4-vision-preview, gpt-4-1106-preview, gpt-4-0613, gpt-3.5-turbo-0613, and davinci-002. The gpt-4-vision-preview model received color stimuli as images, while other GPT-3 and GPT-4 models received RGB code values to specify colors (e.g., [0, 0, 0] for Black) (Table 1). The gpt-4o-2024-05-13 model was tested under two conditions: using visual images (denoted as "gpt4o_visual") and using RGB values (denoted as "gpt4o_text"). Thus, there are seven conditions to evaluate LLMs in total (Table 3).

LLMs received the same 136 Japanese questions as human participants (17 colors × 8 categories; Table 2) via the OpenAI Python API— 'chat.completions' for all GPT-4/3.5 models and 'completions' for *davinci-002*. Each question was posed 100 times with the ten answer choices randomly shuffled, yielding 13 600 prompts per model. Temperature was fixed at 0.0 so that a model always returned its highest-probability option, removing sampling noise and making runs reproducible. Latent variability was still captured: when two choices carried identical top probabilities, the model's internal tie-breaker selected whichever token appeared first, and the changing list order exposed those ties in the response histograms. Hence the reported entropies reflect how sharply each model concentrates probability mass rather than an artefact of stochastic sampling. Trials in which a model produced no answer or multiple answers were discarded; the proportion of valid trials was 100 % (*gpt4o_visual*), 99.99 % (*gpt4o_text*), 99.10 % (*gpt4_vision*), 99.97 % (*gpt4_1106*), 99.59 % (*gpt4_0613*), 99.49 % (*gpt-3.5_turbo-0613*), and 96.61 % (*davinci-002*). All analyses were performed on these valid trials.

| Name | Model | GPT Version | Release Date | Color Input |
|---|---|---|---|---|
| **gpt4o_vision** | gpt-4o-2024-05-13 | GPT-4o | May 13, 2024 | Image |
| **gpt4o_text** | gpt-4o-2024-05-13 | GPT-4o | May 13, 2024 | RGB code |
| **gpt4_vision** | gpt-4-vision-preview | GPT-4 | November 6, 2023 | Image |
| **gpt4_1106** | gpt-4-1106-preview | GPT-4 | November 6, 2023 | RGB code |
| **gpt4_0613** | gpt-4-0613 | GPT-4 | June 13, 2023 | RGB code |
| **gpt35_turbo_0613** | gpt-3.5-turbo-0613 | GPT-3.5 | June 13, 2023 | RGB code |
| **davinci-002** | davinci-002 | GPT-3 | March 15, 2022 | RGB code |

**Table 3: Conditions used to evaluate LLMs.** Their versions, release dates, and color input methods are listed. We use names specified in "Name" columns to report results throughout this paper.

*C. Data Analysis*

Custom Python scripts were used for all analyses. We calculated histograms to quantify word selection frequencies for each color-question combination in both human and LLM data. Results were summarized as heatmaps for each model and question category. In total, 64 heatmaps were generated (8 conditions (7 LLM conditions + human data) × 8 word categories) (Figures 1 and 2). Entropy for a distribution for a given combination of color and category was calculated using the entropy function from the SciPy library with the natural logarithm (i.e. the unit is [nats]; Figure 3).

To assess LLM performance in predicting the most frequently chosen word by human participant, we calculated the percentage of correct choices within valid trials for each LLM model across 8 categories and 17 colors. The chance level was 10% due to the 10 choices per question (Figures 4, 5, and 6). For Figure 4 we report the median accuracy per model together with its 95 % confidence interval, obtained via a percentile bootstrap with 10,000 resamples.

To quantify performance differences simultaneously across models, word categories, and stimulus colors we fitted a cluster-bootstrap ridge-logit model that contains both interactions:

*Correct ~C(Model)+ C(Category) + C(Color)+ C(Model)\*C(Category) + C(Model)\*C(Color)*

Here *Correct* is a trial-level binary outcome that equals 1 when the LLM selected the human most-frequent word for that color-category pair and 0 otherwise. Predictors are declared with *C(·)* in *patsy/statsmodels*, which treats a factor as categorical and expands it into dummy variables. The factors are Model (7 LLM conditions), Category (8 word categories), and Color (17 colors). A mild L2 ridge penalty ($\alpha = 0.1$) was added to stabilize estimation under quasi-separation. We adopted this cluster-bootstrap approach to obtain stable, robust covariance estimates. Parameter uncertainty was obtained through a cluster bootstrap with B = 2000 resamples, clustering on Item (unique color × category trials). Block-wise Wald $\chi^2$ tests, computed from the bootstrap covariance matrices, were used to evaluate the two interactions (Model × Category, Model × Color) and the three main effects (Model, Category, Color). Hereafter, we refer to this model as *the regression analysis*. We did not include the Category × Color interaction because the present analysis is aimed specifically at how performance changes across model generations.

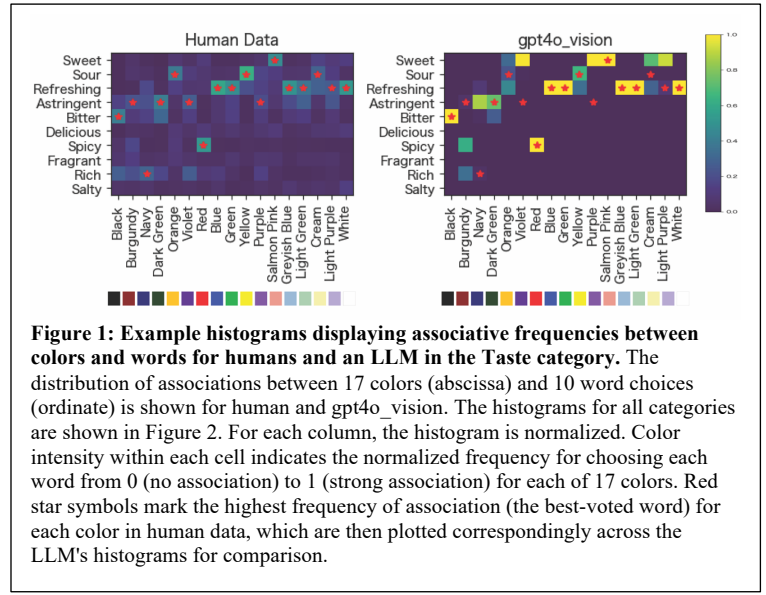

**Figure 1: Example histograms displaying associative frequencies between colors and words for humans and an LLM in the Taste category.** The distribution of associations between 17 colors (abscissa) and 10 word choices (ordinate) is shown for human and gpt4o_vision. The histograms for all categories are shown in Figure 2. For each column, the histogram is normalized. Color intensity within each cell indicates the normalized frequency for choosing each word from 0 (no association) to 1 (strong association) for each of 17 colors. Red star symbols mark the highest frequency of association (the best-voted word) for each color in human data, which are then plotted correspondingly across the LLM's histograms for comparison.

To compare color discrimination abilities between humans and LLMs, we created combined histograms concatenating data from 8 word categories for each color (80-dimensional vectors) (Figure 7A). Pairwise Jensen-Shannon divergence was calculated for all combinations of the 17 colors, resulting in 17×17 matrices for human and LLM data. These divergence values were visualized as heatmaps to represent overall color discrimination patterns (Figures 7B and 7C).

To compare LLM performance with human data, we created scatter plots of human versus LLM divergence values for each model (Figure 7D). Pearson correlation coefficients between normalized human and LLM distance matrices were calculated to quantify the similarity of color discrimination patterns. Fisher transformations were applied to obtain 95% confidence intervals for these correlation estimates (Figure 6E).

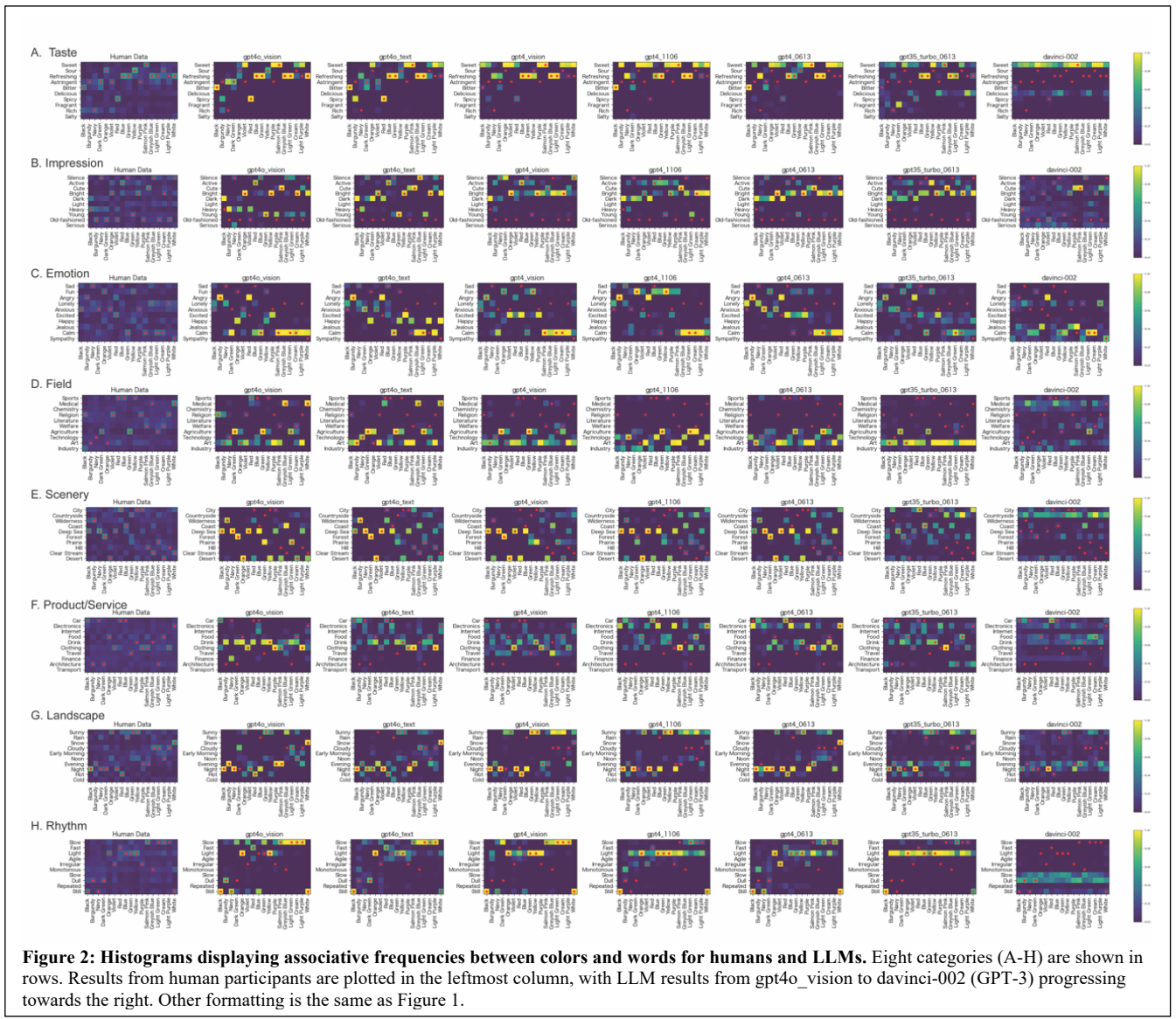

**Figure 2: Histograms displaying associative frequencies between colors and words for humans and LLMs.** Eight categories (A-H) are shown in rows. Results from human participants are plotted in the leftmost column, with LLM results from gpt4o_vision to davinci-002 (GPT-3) progressing towards the right. Other formatting is the same as Figure 1.

## RESULTS

We first plotted the frequency of chosen words for each color in histograms of color-word associations across eight categories (Taste, Impression, Emotion, Field, Scenery, Product/Service, Landscape, and Rhythm) for both human participants and various LLM models (Figures 1 and 2). The histograms reveal distinct patterns of association between colors and words, with notable variations among human and LLMs. We describe further qualitative and quantitative description of results in the following.

### A. Distribution flatness

Overall, human responses show more distributed associations across the color-word pairs, indicated by a broader spread of color intensities (Figure 2, the leftmost column). In contrast, LLM responses, particularly for more advanced models like GPT-4, tend to show more concentrated associations, as evidenced by the brighter, more focused areas in their respective histograms. To quantify this difference in distribution flatness, we calculated the entropy of histograms for human and LLM data (Figure 3). Human responses demonstrated the highest median entropy, approximately 1.9 [nats], and the narrow interquartile range, indicative of greater variability in color-word associations among participants for most of categories and colors. In contrast, LLMs generally exhibited lower entropy values, suggesting more focused or deterministic association patterns. Also, the interquartile range is wider than that for human data, indicating that there is more variability in the degree of uncertainty across categories and colors. A general trend of decreasing entropy from earlier to more recent LLM versions was observed, with GPT-4 models showing lower entropy compared to GPT-3.5 or GPT-3, potentially indicating an evolution towards more consistent color-word associations in newer LLM iterations. The GPT3 model (davinci-002) exhibited the highest median entropy and the widest range of entropy values.

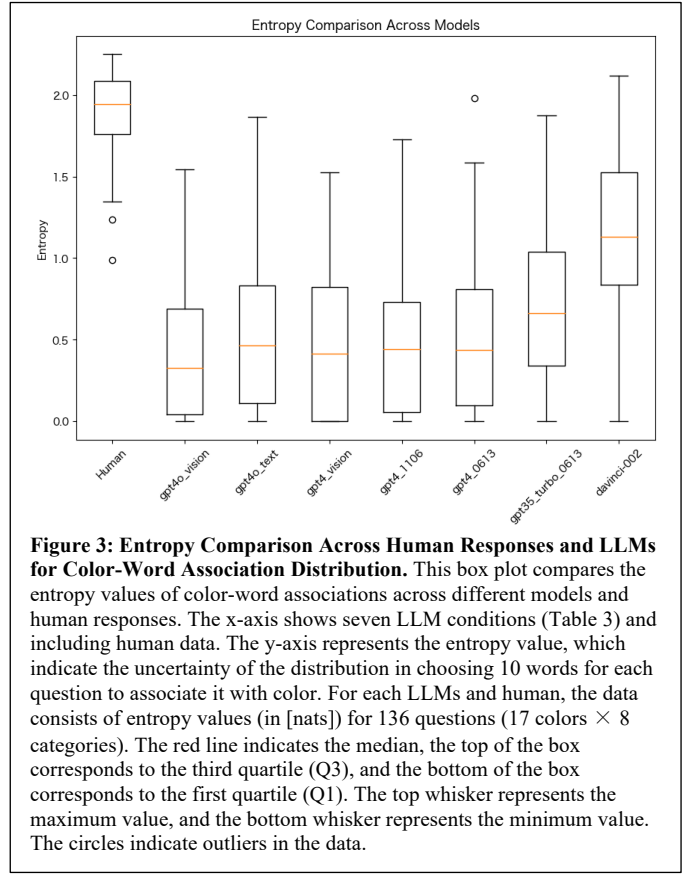

**Figure 3: Entropy Comparison Across Human Responses and LLMs for Color-Word Association Distribution.** This box plot compares the entropy values of color-word associations across different models and human responses. The x-axis shows seven LLM conditions (Table 3) and including human data. The y-axis represents the entropy value, which indicate the uncertainty of the distribution in choosing 10 words for each question to associate it with color. For each LLMs and human, the data consists of entropy values (in [nats]) for 136 questions (17 colors × 8 categories). The red line indicates the median, the top of the box corresponds to the third quartile (Q3), and the bottom of the box corresponds to the first quartile (Q1). The top whisker represents the maximum value, and the bottom whisker represents the minimum value. The circles indicate outliers in the data.

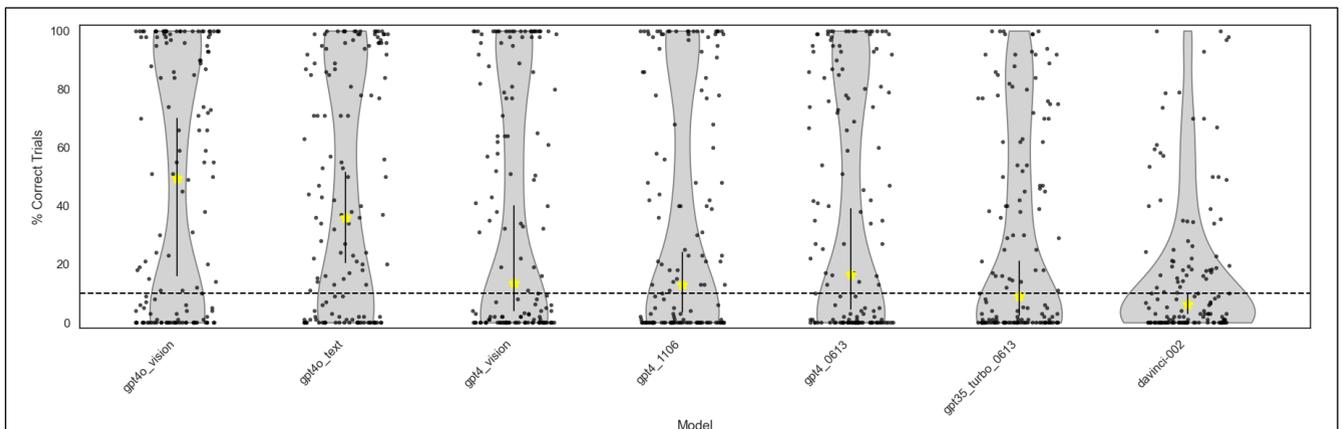

**Figure 4. Distribution of percentage correct responses across LLMs in color-word association tasks.** For every model (abscissa) the violin shows the distribution of percentage-correct scores obtained for the 136 color-category items (17 colors × 8 categories); wider sections indicate a higher density of items at that accuracy. Black dots are the individual item scores, jittered horizontally to avoid overlap. The yellow star marks the sample median, and the vertical black error bar attached to each star is the 95 % percentile-bootstrap confidence interval for that median (10,000 resamples within model). The dashed horizontal line at 10 % denotes chance performance given ten response options.

## B. Newer LLMs predict better human color-word association.

The performance of LLMs in color-word association reveals a clear trend of improvement from older to newer models (Figure 4). Across the seven test conditions the cluster-bootstrap 95 % confidence intervals (CIs) for the median percentage of correct trials were gpt4o-vision 16.0 – 70.0 %, gpt4o-text 20.5 – 51.5 %, gpt4-vision 4.0 – 40.0 %, gpt4-1106 3.5 – 24.0 %, gpt4-0613 4.5 – 39.0 %, gpt-3.5-turbo-0613 2.5 – 21.0 %, and davinci-002 3.0 – 10.1 %. Among these, gpt4o-vision attains the highest central accuracy and a response distribution whose upper half frequently exceeds 80 % correct, while gpt4o-text occupies an intermediate position with its CI extending into the low-50 % range. gpt4-vision and gpt4-1106 achieve more modest median levels but still place many individual color-category items above 80 %. In contrast, the earlier gpt-3.5-turbo-0613 and especially davinci-002 cluster near chance (10 %), and their CIs never reach beyond the low-twenties.

The regression analysis further revealed significant main effects for all three factors: Model ($\chi^2$ (144) = 227.0, $p < 0.001$), Category ($\chi^2$ (49) = 99.9, $p = 2.5 \times 10^{-5}$), and Color ($\chi^2$ (112) = 211.6, $p = 4.0 \times 10^{-8}$). Thus, overall accuracy differs systematically across LLM generations, across semantic word categories, and across stimulus hues. Examining interactions, the Model × Category term remained significant ($\chi^2$ (42) = 71.6, $p = 0.003$), indicating that the size of the model improvement varies by word categories. By contrast, the Model × Color interaction was not reliable ($\chi^2$ (96) = 96.7, $p = 0.46$), suggesting that color-specific effects are largely additive rather than model-dependent. We next dissect performance by word category—and subsequently by color—to illustrate in detail how these main-effect and interaction patterns play out across the models.

## C. Difference across word category

To evaluate variations across different word categories, we created violin plots summarizing the prediction performance of LLMs according to word categories (Figure 5). Overall, there is a clear performance gradient from newer to older models, with davinci-002 consistently showing the lowest accuracy across categories. GPT-4 models, particularly gpt4o_vision and gpt4o_text, mostly outperform older models across all categories, with their maximum accuracy indicated in the upper part of the violin plot often reaching up to 100% in some cases. In the Taste category, gpt4o showed greater performance than previous LLMs as indicated by the median performance value above 70% in gpt4o_vision. The Field category also similarly demonstrated high performance specifically by gpt4o_vision, but not gpt4o_text. The Landscape and Rhythm categories maintain a high level of performance across GPT-4 models overall. In contrast, the Impression and Scenery categories present a significant challenge for all models. In the Emotion category, the gpt4_vision model showed the highest median performance, while GPT-4o showed lower performance. The regression analysis confirmed a strong main effect of Category ($p < 0.001$), demonstrating that some semantic domains are intrinsically easier for all models. Complementing this finding, the significant Model × Category interaction ($p = 0.003$) shows that the size of the performance gap between newer and older models differs across semantic domains.

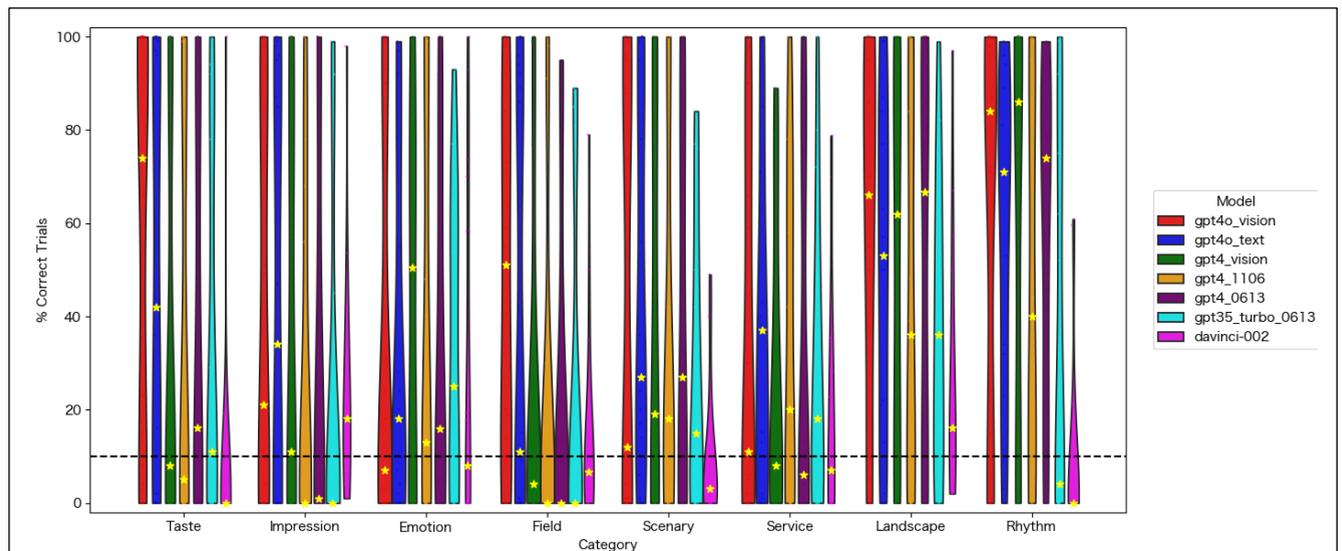

**Figure 5. Distribution of percentage correct responses across categories and LLMs in color-word association tasks.** Each bar represents the percentage of correct trials for a specific LLM condition within each word category. Word categories are shown along the abscissa, and the percentage of correct responses is displayed on the ordinate. Seven LLM conditions are distinguished by different colors as indicated in the legend. Yellow asterisks indicate the median percentage correct for each LLM-category combination. A horizontal dashed line at 10% marks the chance level performance. Here, Product/Service category is denoted as "Service" for brevity.

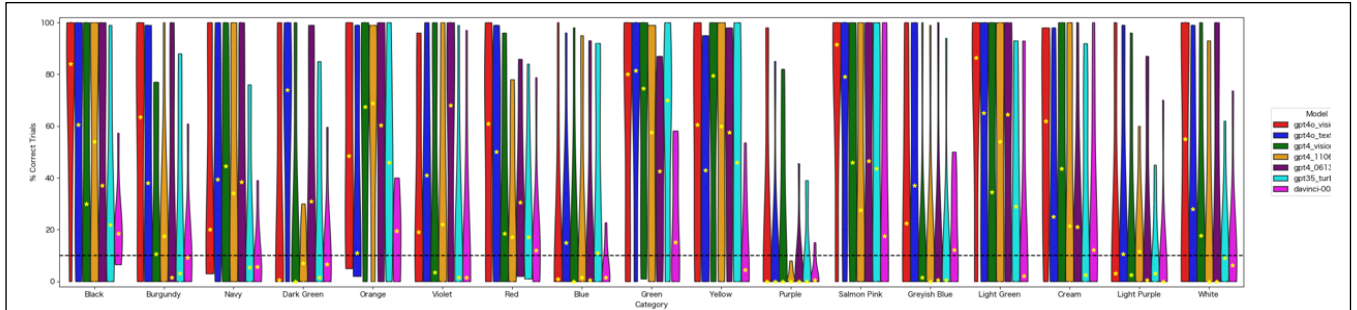

**Figure 6. Distribution of percentage correct responses across colors and LLMs in color-word association tasks.** Each bar represents the percentage of correct trials for each LLM condition for each color. Colors are shown along the abscissa, and the percentage of correct responses is displayed on the ordinate. Seven LLM conditions are distinguished by different colors as indicated in the legend. Yellow asterisks indicate the median percentage correct for each LLM-color combination. A horizontal dashed line at 10% marks the chance level performance.

### D. Difference across colors

Next, we plotted the performance across different colors to examine variability in LLMs' performance (Figure 6). The regression analysis revealed a strong main effect of Color ($p = 4 \times 10^{-8}$), confirming that certain colors are intrinsically harder for every model. The GPT-3 (davinci-002) model consistently underperformed across all colors, often barely surpassing the chance level. Newer models generally performed better, but interestingly, we found the performance varied significantly across colors. In particular, cool-toned colors such as Purple, Blue, Greyish Blue, and Light Purple presented greater challenges, with performance near or below chance levels across all models. On the other hand, colors like Black, Burgundy, Red, and White showed marked increases in performance for GPT-4o. Cream was particularly better for gpt4o_vision, but not in the other models including gpt4o_text. Conversely, Navy, Dark Green, and Violet demonstrated high performance for gpt4o_text but not gpt4o_vision. Despite these descriptive differences, the Model × Color interaction in the regression analysis did not reach statistical significance ($p = 0.46$), indicating that color-specific variations are largely captured by the overall main effects of Model and Color rather than by their interaction.

As we reviewed the overall trends and differences across word categories, several notable observations emerged for particular cases as follows.

  a) **Taste category**

In the Taste category, strong associations between certain colors and tastes were evident (Figure 2A). Human participants strongly associated Red with "spicy," Yellow with "sour," and Black with "bitter." GPT-4o captured these associations particularly well. For instance, the strong Red-Spicy and Yellow-Sour associations were evident in GPT-4o, but not so much in older models. Older versions of LLMs tended to associate red and yellow with "sweet." Regardless of color, older models tended to show a bias towards choosing "sweet," most prominently seen in davinci-002.

  b) **Impression category**

The Impression category revealed interesting disparities between human responses and LLM predictions (Figure 2B). Humans associated "old fashioned" with many colors (Burgundy, Dark Green, Violet, Purple, Light Green, Light Purple), but this association was rarely chosen by any version of LLM. Overall, GP-4o showed better performance than previous versions of LLM. In particular, humans associated green with "young" most frequently, and GPT-4o (gpt4o_vision and gpt4o_text) replicated this association, but not older versions of LLM.

  c) **Emotion category**

In the Emotion category, human responses showed associations with Red strongly linked equally to "angry" and "excited," while Blue correlated with "sad" (Figure 2C). LLMs tended to relate Red with "anger" more strongly than humans, although only gpt4_vision captured humans' balanced association to "anger" and "excited" very well. LLMs often associated Blue with "calm" and "excited," and no LLM showed a peak response at "sad" with Blue as in humans. Black presented another interesting example of differences between humans and LLMs. Humans related Black with "anxious" most strongly, but LLMs tended to associate it more with "lonely," although "anxious" was chosen equally in gpt4o_vision, gpt4_vision, and gpt35_turbo.

  d) **Field category**

In the Field category, human responses showed clear associations between certain colors and fields (Figure 2D). For example, green was strongly associated with "agriculture," blue with "sports," and white with "medicine." LLMs generally captured these associations well for green and white, but they tended to associate blue with "art" or "technology."

*e) Scenery category*

The Scenery category demonstrated clear matches between human and LLM responses for dark colors such as black, burgundy, navy, and dark green (Figure 2E). Blue was heavily associated with "sea" and "sky," green with "forest" and "grassland," and yellow with "desert" for most LLMs. However, davinci-002 showed a biased response towards "countryside" for most colors.

*f) Product/Service category*

In the Product/Service category, human responses showed relatively strong associations between "clothing" and Violet, Purple, Salmon Pink, and Light Purple (Figure 2F). These associations were well predicted by gpt4o_vision, gpt4o_text, and gpt4_vision, but older models missed them for some of these colors. Interestingly, GPT-4 models with visual inputs had difficulty predicting humans' association of Red with "car," but other LLMs with text inputs (except GPT-3.5 turbo) correctly predicted this association.

*g) Landscape category*

For the Landscape category, the overall performance in predicting human responses was high in most LLMs except davinci-002, which showed biased responses towards "Night" and "evening" (Figure 2G). Gpt4o_vision showed particularly good alignment with human responses for colors except Navy, Blue, Greyish Blue, and Light Purple.

*h) Rhythm category*

The Rhythm category also demonstrated an overall high level of median performance across many LLM conditions. However, some LLM conditions showed biased responses to particular words, such as "light" (gpt4_1106, gpt4_0613, gpt35_turbo), and "slow" and "dull" (davinci-002) for most colors (Figure 2H). "Light" was indeed the best voted word by humans for four colors (orange, blue, green, yellow), and thus this bias happened to contribute to better performance overall. In contrast, GPT-4o did not show such clear bias and achieved high performance in predicting the best human-voted word overall. The gpt4_vision condition accurately predicted the best word "dull" chosen by humans, but this was the only case apart from davinci-002 showing a bias for choosing "dull" for all colors.

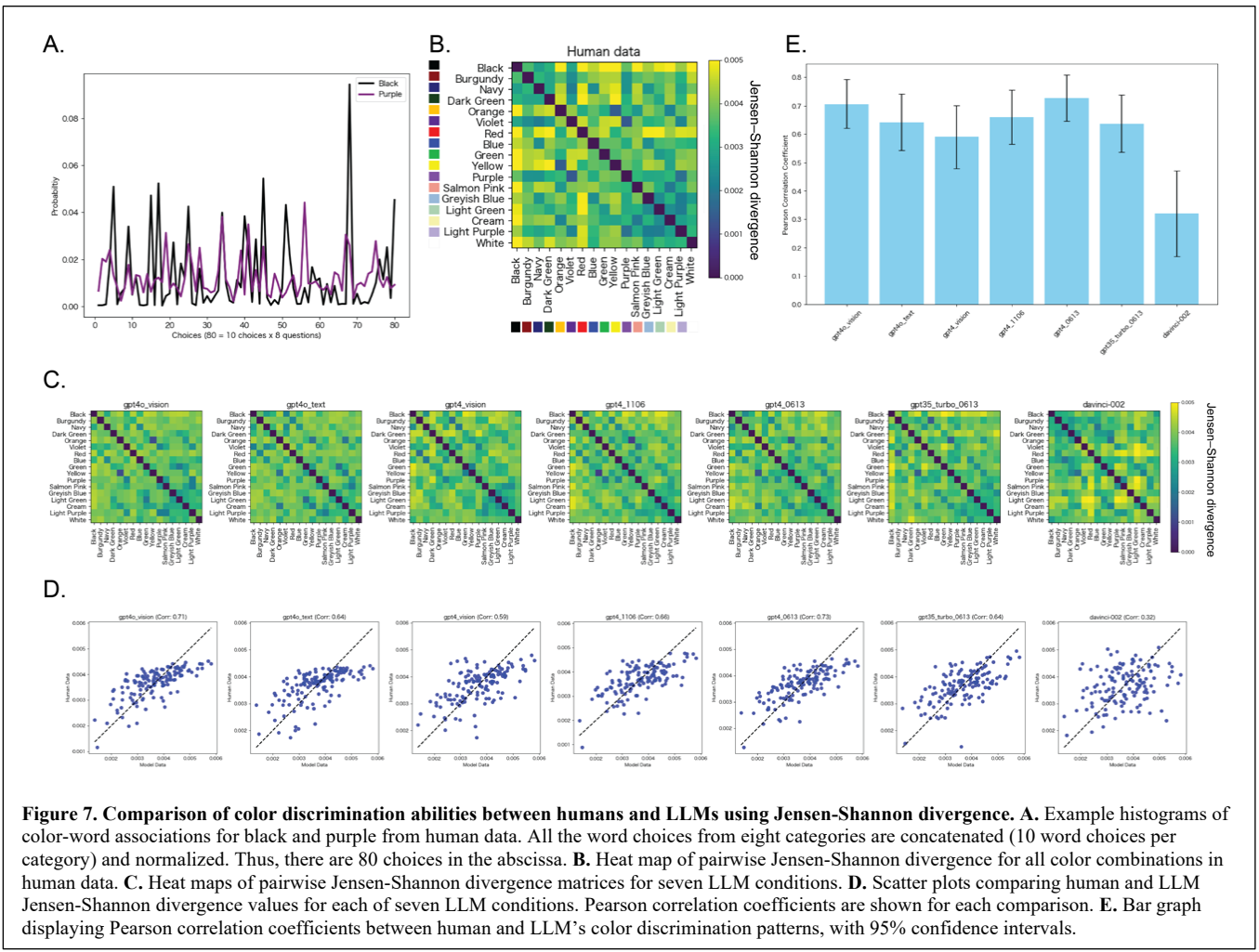

**Figure 7. Comparison of color discrimination abilities between humans and LLMs using Jensen-Shannon divergence. A.** Example histograms of color-word associations for black and purple from human data. All the word choices from eight categories are concatenated (10 word choices per category) and normalized. Thus, there are 80 choices in the abscissa. **B.** Heat map of pairwise Jensen-Shannon divergence for all color combinations in human data. **C.** Heat maps of pairwise Jensen-Shannon divergence matrices for seven LLM conditions. **D.** Scatter plots comparing human and LLM Jensen-Shannon divergence values for each of seven LLM conditions. Pearson correlation coefficients are shown for each comparison. **E.** Bar graph displaying Pearson correlation coefficients between human and LLM's color discrimination patterns, with 95% confidence intervals.

*E. Color discrimination*
As previous studies demonstrated LLMs' ability to discriminate colors in a manner similar to humans [11,12], here we quantify and compare the color discrimination abilities of humans and LLMs, based on difference in color-word associations across colors. To do so, we calculated pairwise Jensen-Shannon divergence that quantifies similarity between two histograms of color-word associations (Figure 7A) for all possible color combinations, represented as a heatmap matrix (Figure 7B, see Methods). The human divergence matrix (Figure7B) revealed distinct patterns of color discrimination, with certain color pairs showing higher divergence, indicating greater perceived differences among colors. For example, the divergences within dark colors such as black, burgundy, navy, and dark green showed lower values, as indicated in the upper right corner of the matrix. In contrast, there are noticeable differences from those calculated from LLM data (Figure 7C). We observed varying degrees of similarity to human data across models. The scatter plots (Figure 7D) illustrate the relationship between human and LLM divergence values, with correlation coefficients ranging from 0.32 (davinci-002) to 0.73 (gpt4_0613). The bar graph of Pearson correlation coefficients shows that the 95% confidence intervals are overlapping across LLMs except for davinci-002, which exhibited the weakest correlation with human color discrimination patterns (Figure 7E). Overall, the estimated correlation coefficients for GPT-4 and GPT-3.5 models reached around 0.7, suggesting that these LLMs have color discrimination abilities similar to that by humans, even based on differences in color-word associations across different colors.

## DISCUSSION
Our study evaluated the ability of Large Language Models (LLMs) to replicate human color-word associations, revealing both promising advancements and notable limitations. By comparing responses from various LLMs from GPT-3 to GPT-4o with both text and visual inputs—to data from over 10,000 Japanese participants, we aimed to assess how well these models emulate human cognitive processes in associating colors with words.

*A. Distribution flatness and relevance to diversity*
The observed trend of decreasing entropy from earlier to more recent LLM versions suggests an evolution towards more consistent and focused color-word associations (Figure 3). This progression may reflect improvements in training data quality, model architecture, or fine-tuning processes. However, it also raises questions about the trade-off between consistency and the ability to capture the full spectrum of human associative diversity. GPT-3 was previously shown to generate basic color terms at a frequency similar to that of human non-synaesthetes, with better performance at higher temperature values [10]. This is thought to be due to an increase in the generation of minor color terms. Our data from human participants reflect the diversity of the Japanese population with varied demographics, including gender, age, and geographical locations (Methods). Such diversity could be more relevant to the shape of distribution and thus better captured potentially by changing the temperature value of LLMs.

In the current study, we set the temperature parameter to 0 for all models in this study (Methods). This instructs the model to always choose the most probable next token, effectively eliminating randomness in the output. This relatively deterministic approach explains the lower entropy and more concentrated associations we observed in LLM responses. The diversity in human responses and the temperature parameter setting in LLMs are also reasons we evaluated the best-voted word, the peak of the distribution, to assess LLMs' performance in predicting color-word associations. However, because this top-1 criterion credits only the single most frequent human response, items that have two or more comparably prominent human answers (e.g., Red being linked to both 'spicy' and 'sweet') may yield conservative estimates of model skill. Our preliminary observations confirm that the peak location of the distribution from LLMs' choices was not affected much by the temperature parameter. Future studies could probe response variability by manipulating instructional prompts that ask an LLM to 'speak as' individuals with specific attributes—such as gender, profession, or educational background—an approach already shown to introduce systematic biases [17]. Examining how those biases shift color–word associations would clarify whether LLMs can capture the differences observed across people with diverse attributes.

*B. Improvement in LLM Performance Across Generations*

Our findings reveal a clear progression in the performance of LLMs, with newer models demonstrating improved accuracy in predicting human best-voted color-word associations. The GPT-4 model with visual inputs exhibited a median accuracy rate exceeding 50%, and around 40% even with text inputs to specify colors. This is notable given that previous GPT-4 models with visual or text inputs demonstrated a median accuracy of only around 20%. Furthermore, although the median accuracies are similar between GPT-3.5 and GPT-3 in our results, the lower portion of the violin plot for those models (gpt-35_turbo and davinci-002) is widened, suggesting an increase in lower performance values (Figure 4). This could be expected due to GPT-4's better performance on various academic and professional exams as well as its multimodal capability [13]. It is still noteworthy that this improvement for newer LLMs in color-word association is observed incidentally, as LLMs are not trained to solve color-word associations explicitly. This is consistent with the idea that contemporary LLMs have the potential to capture complex patterns in human cognition compared to their predecessors.

*C. Performance Across Categories and Colors*

The challenges posed by certain categories and colors indicate potential gaps in the models' training data or limitations in their ability to capture more subtle associations. This variability across categories and colors provides valuable insights into the strengths and weaknesses of current LLM architectures in processing and generating human-like perceptual associations.

For example, our human data from the Japanese population associate red with spicy. Such red-spicy association is also reported for non-Japanese populations14, although red is often associated with sweet as well [15]. This makes sense, as various spicy and sweet foods are colored red (e.g., red hot chili pepper, strawberry). In our results, davinci-002 (GPT-3) and gpt4_1106 associated red exclusively with sweet, while gpt3.5_turbo, gpt4_0613, and gpt4_vision associated it with both sweet and spicy. In contrast, GPT-4o (gpt4o_vision and gpt4o_text) chose spicy exclusively, which corresponds to the best-voted word by humans. This example illustrates the evolution of LLMs' ability to capture nuanced and culturally relevant color-word associations, with the latest models showing improved alignment with human preferences.

On the other hand, the relatively low performance in Impression and Emotion categories might reflect LLMs' difficulty in associating abstract or subjective words with colors. For example, there were discrepancies in associations for "old fashioned" between humans and LLMs in the Impression category. The Emotion category highlighted varying success in replicating human associations, such as red with both "angry" and "excited", or blue with "sad". These examples could be considered as abstract or subjective descriptions of colors. In fact, a previous study showed that color space alignment with LLMs' description of colors drops considerably in the presence of more abstract or subjective color descriptions in an earlier generation of LLM [16].

*D. Color Discrimination Abilities*

Our analysis of color discrimination abilities shows that LLMs except davinci-002 (GPT3) can differentiate colors in ways similar to humans, even based on colors represented by the degree of associations to various words. The high correlation coefficients (around 0.7) between human and LLM color discrimination patterns suggest that these models have developed differential representation of colors similar to that by humans. On the other hand, previous studies have shown overall high discriminability of colors in LLMs[11,12], with GPT-3's performance on color discrimination being slightly better than GPT-3.5, showing high correlation coefficients around 0.812. This is also parallel to a study where an earlier version of LLM performed well at predicting color comparatives despite their difficulty in color space alignment [16].

Whereas these observations from previous studies are consistent with our results in advanced LLMs such as GPT-4 and GPT-3.5, they are in contrast with our results for GPT-3, which showed a much lower correlation coefficient of approximately 0.3 (Figure 7E). This discrepancy might be explained by differences between the current and previous studies. First, the previous studies had LLMs judge the similarity of a given pair of colors and directly compare it with human similarity judgments [11,12]. Second, GPT-3 generally has poorer capability in handling Japanese compared to English, given that these models are often trained predominantly on English-language data, although LLMs' ability to solve various tasks has also improved substantially in GPT-4 or GPT-4o[13],[18–20]. Third, the use of different metrics to quantify color discrimination could provide different insights. In our study, we used correlation coefficients with Jensen-Shannon divergence to calculate the divergence between color-word association histograms. A recent study shows that another metric to quantify similarity demonstrated better alignment to human color discriminability for GPT-4 than GPT-3.5, which was not found when using correlations to quantify the discriminability [11]. Thus, these suggest the importance of considering both the choice of metric and the language-specific capabilities of LLMs when evaluating their color discrimination abilities.

## Conclusion

Our study demonstrates the evolution of LLMs in capturing human color-word associations, while also highlighting open questions about how their internal representations compare with human semantic memory; clarifying this will require targeted future work, including possibly exploring additional metrics (e.g. top-k accuracy). At the same time, we replicate earlier findings that LLMs can distinguish colors as accurately as people, while demonstrating that the words associated with those colors still diverge in systematic ways. By showing both the convergence in basic color discrimination and the persisting gap in color–word association, we underscore that surface-level perceptual gains may mask deeper representational differences between humans and LLMs—differences future work needs to unpack.

The observed improvements across LLM generations suggest promising applications in fields such as cognitive science, design, and cross-cultural communication in providing insights related to color representation. LLMs could eventually become tools to guide the relationship between colors and words, often estimated based on individual intuition or sensibility, towards one that aligns well with human perception and cognition. This could contribute to improvements in the user experience of design products and enhance the effectiveness and efficiency of marketing strategies based on valid color-word associations. On the other hand, the challenges in color-word association might reflect LLMs' limitations in simulating human cognitive functions. When interpreted in a broader context, this could mean that basic cognitive tasks are useful in evaluating LLMs' ability to align with human capabilities, thereby functioning as a tool to detect potential risks in LLMs. As LLMs continue to advance, it will be crucial to consider both their expanding capabilities and the societal implications of their use. Eventually, LLMs could become well-equipped to simulate human cognitive functions better and, in turn, serve as valuable instruments for studying the mechanisms underlying human cognition on a large scale.

## ACKNOWLEDGEMENT

We would like to thank Takayuki Kitahara, Saki Kanada, and Tomotake Kozu for support.